\documentclass[letterpaper, 10 pt, conference]{ieeeconf}  % Comment this line out if you need a4paper

\IEEEoverridecommandlockouts                              % This command is only needed if 
\let\labelindent\relax

\usepackage{graphics} % for pdf, bitmapped graphics files
\usepackage{amsmath} % assumes amsmath package installed
\usepackage{amssymb}  % assumes amssymb package installed
\usepackage{amsthm}
\usepackage{tikz}
\usepackage[hidelinks]{hyperref}
\usepackage[]{cite}
\usepackage[]{subcaption}
\usepackage{mathnotation}

\title{\LARGE \bf
Optimal Gait Families using Lagrange Multiplier Method
}

\author{Jinwoo Choi, Capprin Bass, and Ross L. Hatton% <-this % stops a space
\thanks{This work was supported in part by the National Science Foundation under awards CMMI-1653220 and 1826446}% <-this % stops a space
\thanks{
    All authors are with the Collaborative Robotics and Intelligent Systems (CoRIS) Institute at Oregon State University, Corvallis, OR USA. 
    {\tt\small <\href{mailto:choijinw@oregonstate.edu}{choijinw},
    \href{mailto:basscap@oregonstate.edu}{basscap},
    \href{mailto:ross.hatton@oregonstate.edu}{Ross.Hatton}>@oregonstate.edu} \tt\small
}
\thanks{(Corresponding Author : Ross L. Hatton)}
}
    
\begin{document}

\theoremstyle{definition}
\newtheorem{theorem}{Theorem}[]

\maketitle
\thispagestyle{plain}
\pagestyle{plain}

%%%%%%%%%%%%%%%%%%%%%%%%%%%%%%%%%%%%%%%%%%%%%%%%%%%%%%%%%%%%%%%%%%%%%%%%%%%%%%%%
\begin{abstract}
The Robotic locomotion community is interested in optimal gaits for control. Based on the optimization criterion, however, there could be a number of possible optimal gaits. For example, the optimal gait for maximizing displacement with respect to cost is quite different from the maximum displacement optimal gait. Beyond these two general optimal gaits, we believe that the optimal gait should deal with various situations for high-resolution of motion planning, e.g., steering the robot or moving in ``baby steps.'' As the step size or steering ratio increases or decreases, the optimal gaits will slightly vary by the geometric relationship and they will form the families of gaits. In this paper, we explored the geometrical framework across these optimal gaits having different step sizes in the family via the \textit{Lagrange multiplier method}. Based on the structure, we suggest an optimal locus generator that solves all related optimal gaits in the family instead of optimizing each gait respectively. By applying the optimal locus generator to two simplified swimmers in drag-dominated environments, we verify the behavior of the optimal locus generator.
\end{abstract}

%%%%%%%%%%%%%%%%%%%%%%%%%%%%%%%%%%%%%%%%%%%%%%%%%%%%%%%%%%%%%%%%%%%%%%%%%%%%%%%%
\section{Introduction}

Mobile robots move by using the interaction between the robot and its environment to convert internal joint motion into motion of the body. For robots whose joint motion is bounded (as compared to the continuous rotation afforded by wheels and propellers), it is often useful to focus attention on gaits---cyclic shape inputs that produce characteristic net displacements and individually remain within the joint limits. These gaits can then be sequenced into motions that are guaranteed to respect the joint limits.

The maximum displacement a system can produce in each gait cycle is typically on the order of the system's body length \cite{hatton_geometric_2011}. Motion plans over long distances are therefore dominated by repetitions of a gait that maximizes speed or efficiency under constraints on effort or time \cite{ramasamy_soap-bubble_2016, tam_optimal_2007}. Much locomotion research thus focuses on identifying or understanding such gaits. A perspective that we have found useful \cite{ramasamy_geometric_2017} is that these gaits lie at points in the control space where the gradient of induced displacement with respect to the gait parameters is in equilibrium with the gradient of the effort function.

\begin{figure}
    \centering
    \includegraphics[width=0.48\textwidth]{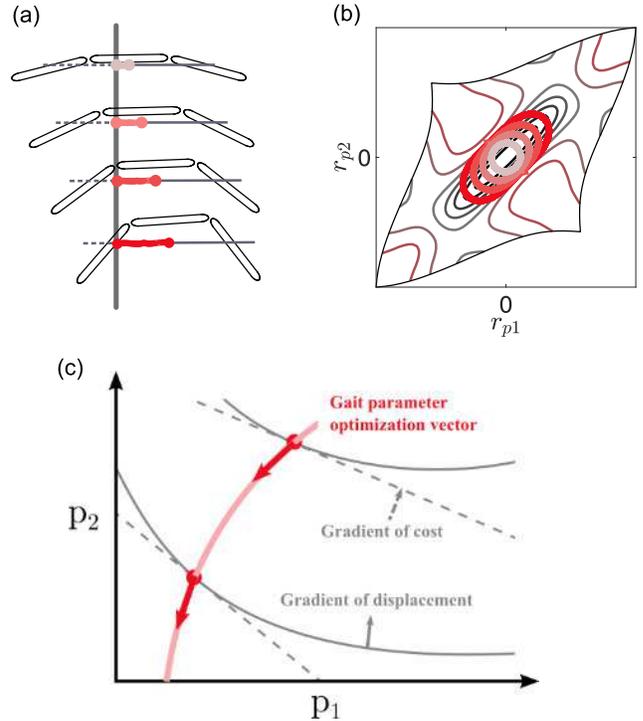}
    \caption{(a) Step-optimal gait trajectories for the drag-dominated three-link swimmer. Based on the constrained displacement, the different behavior of the robots can be identified: Robots try to minimize their joint motions for accomplishing the given displacement. (b) Step-optimal gaits in natural metric coordinates. Gait size decreases with displacement. (c) The geometric meaning of Lagrange multipliers is that the gradient of the cost and constraint functions are parallel to each other at the optimal point. To generate an optimal trajectory, the gait parameter optimization vector is needed.}
    \label{fig:StepOptimalMotion}
\end{figure}

Beyond maximum-speed or maximum-efficiency gaits for crossing large distances, motion plans must often incorporate gaits that are optimized for different criteria, e.g., steering the system or taking ``baby steps" that---although less efficient than the system's ``natural stride"---allow for finer-grained motion plans. These gaits correspond to the solutions of constrained optimization problems in which the constraints are defined on the per-cycle displacement, e.g. the magnitude of net displacement for ``baby steps" as illustrated in Fig.~\ref{fig:StepOptimalMotion}(a-b), or the ratio between net rotation and net translation for a specified steering rate.

For any given displacement constraint, it is relatively straightforward to directly optimize for a gait that maximizes some notion of speed or efficiency while satisfying the constraint \cite{hatton2021inertia}. Families of gaits---e.g., related by different step sizes or steering rates---can then be generated by individually optimizing over a sampling of constraint values. To the best of our knowledge, however, there has been little to no work on the relationship between these ``step-optimal" gaits across changes in the generating constraint.

In this paper, we identify the mathematical structure underlying the step-optimal gait optimization problem, and present a process for generating families of step-optimal gaits without individually optimizing each gait. From our proposed perspective, step-optimal gaits can be seen as points in control space where the level sets of the gait cost and constraint functions are tangent to each other (such that their gradients with respect to the gait parameters are parallel). Curves passing through these osculation points, as illustrated in Fig.~\ref{fig:StepOptimalMotion}(c), contain sets of step-optimal gaits, and can be naturally parametrized by the values of the generating constraint function.

Formally, the location of step-optimal gaits at points where the gradients of the cost and constraint functions are parallel is a classic Lagrange-multiplier solution to a constrained optimization problem.\cite{chong2004introduction} To generate the locus of step-optimal gaits across constraint values, we take a derivative of the Lagrange-multiplier structure, producing a Hessian matrix for the given optimality condition.\cite{fletcher2013practical} Given a starting point on the step-optimal curve (which we can find by optimizing for a single value of the constraint function), we generate a step-optimal curve by projecting the gradient of the constraint function onto the null-space of the Hessian, and then flowing in the direction of the resulting vector.

To demonstrate our approach, we generate families of step-optimal forward-displacement gaits for three- and four-link swimmers immersed in viscous fluids. The three-link swimmer is a standard reference system in locomotion, and its dynamics have two properties that make the swimmer particularly useful for illustrating the Lagrange-Hessian technique: the net displacement induced by a gait corresponds to the amount of \emph{constraint curvature} that the gait encloses, and the time-energy cost of executing a gait corresponds to its metric-weighted pathlength through the shape space\cite{ramasamy_geometry_2019}. As illustrated in Fig.~\ref{fig:StepOptimalMotion}, the step-optimal gaits found through the Lagrange-Hessian approach are thus solutions to the weighted isoareal/isoperimetric problem---a set of concentric circles stretched along the ``rich" axis of the shape space.

\section{Gait Optimization}

\subsection{Unconstrained Gait Optimization}
For the purposes of our analysis in this paper, a gait is a cyclic trajectory $\gait$ in the shape of a system, e.g., an oscillation of its joint angles. When a system executes a gait, its interactions with its environment generate a characteristic displacement $\fiber_{\gait}$ over each cycle, and the system incurs an associated cost $\alnth_{\gait}$. Displacement may be along one or more directions; in this paper, we assume for simplicity of notation (and without loss of generality) that displacement can be projected down to a single ``interesting" direction. Common costs include some notion of opportunity (e.g., how long does it take to execute the gait?) and effort (e.g., how much energy must the system expend to execute the gait?).

For a given notion of cost and a space of candidate gaits~$\allgaits$, an optimal gait for maximizing the speed or efficiency of a system is one which maximizes the ratio between the induced displacement and the incurred cost,
\begin{equation}
    \gait_{\text{opt}} = \argmax_{\gait}\,\frac{\fiber_{\gait}}{\alnth_{\gait}}.
    \label{eq:optgaitdef}
\end{equation}
If we parametrize $\allgaits$ with a set of parameters $p$, then taking the derivative of the righthand side of~\eqref{eq:optgaitdef} with respect to $p$ gives the first-order condition for gait optimality as
\begin{equation}
    \nabla_{p} \fiber_{\gait} - \frac{\fiber_{\gait}}{\alnth_{\gait}} \nabla_{p} \alnth_{\gait} = \mathbf{0}, \label{eq:unconstrainedflow}
\end{equation}
i.e., an optimal gait is one for which any gains in displacement that can be achieved by varying the parameters are in equilibrium with the extra costs incurred. Note that the lefthand side of~\eqref{eq:unconstrainedflow} defines a vector field which can be flowed along to find a solution to~\eqref{eq:unconstrainedflow}.

\begin{figure*}[ht]
    \centering
    \includegraphics[width=\textwidth]{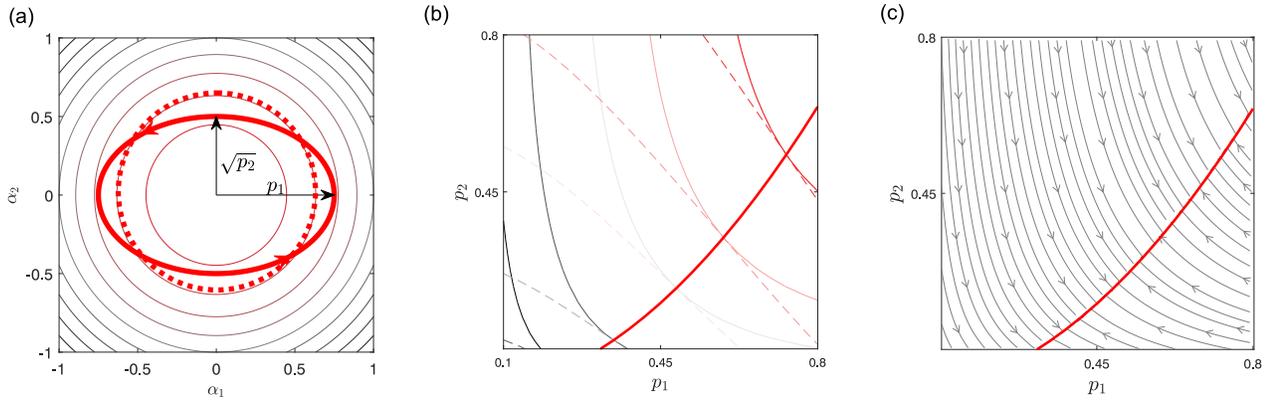}
    \caption{Example system for explaining the geometric intuition for a Lagrange multiplier method. (a) Constraint curvature underlying contours is expressed by a circular paraboloid $(1-\alpha_1^2-\alpha_2^2)$. To describe an elliptical gait, two gait parameters are defined: $p_1$ is the semi-major axis of the  elliptical gait, and $p_2$ is the semi-minor axis square. Then, the optimal parameters make the elliptical gait a circular gait, denoted by the red dotted line. (b)  $p^*$ is a trajectory of optimal parameters for any given displacement. The red line in Fig (b) is a parameter locus for the optimal gait. The set of the dashed lines represents a level set of a cost and the set of the solid lines represent a level set of a net displacement. (c) The vector field is generated by the negative gradient of the Lagrangian function with given $\lambda^*$. The gray line represents the vector field. The stationary line of the Lagrangian Function is illustrated by the red line. Any point on the stationary line is the set of step-optimal gait parameters corresponding to the specific displacement level.}
    \label{fig:LagrangeExample}
\end{figure*}

\emph{Example: Weighted area-perimeter problem.}
For many systems, induced displacement increases with amplitude (bigger shape changes push the system further), but at a diminishing rate (leverage against the environment decreases at extreme angles, so the system gets less displacement per amount of shape change). By a similar token, incurred cost tends to increase monotonically with gait amplitude (large cycles take more time to execute at a given shape velocity, or take more shape velocity---and thus power---to execute in a given time).

The structure of such gait optimization problems is thus qualitatively similar to a weighted area-perimeter problem in which the goal is to enclose as much ``rich" area as possible in a field whose quality diminishes with radius, while minimizing the perimeter of the encircling curve. To intuitively understand the gait optimization dynamics, we can therefore construct a toy problem of this form and use it to generate simple plots that highlight features of the optimization process.

For instance, if we take the ``field quality" as being $q = (1-\alpha_1^2-\alpha_2^2)$, as illustrated in Fig~\ref{fig:LagrangeExample}(a), the closed curve maximizing the ratio of enclosed quality to perimeter is the dashed circle, which lies at the point where the diminishing quality of the field is balanced against the cost of expanding the circle to cost more area.

\subsection{Constrained gait optimization}

While maximum-speed and maximum-displacement gaits are important for moving a system over long distances, a gait-based planner may also need access to gaits for which the induced displacement is smaller than that of the gait satisfying~\eqref{eq:optgaitdef}, e.g., to allow for following specified paths with more accuracy, or for station-keeping without shape drift. These gaits, which we call \emph{step-optimal}, are solutions to the optimization problem
\begin{equation}
    \gait_{\text{s-opt}} = \argmax_{\gait}\,\frac{\fiber_{\gait}}{\alnth_{\gait}} \ \ |\ \ \fiber_{\gait} = g_c,
    \label{eq:soptgaitdef}
\end{equation}
where $g_c$ is a specified induced displacement.

To find sets of gait parameters satisfying this condition, we first note that because $\fiber_{\gait}$ is fixed,~\eqref{eq:soptgaitdef} reduces to the problem of finding the minimum-cost gait for a given displacement,
\begin{equation}
    \gait_{\text{s-opt}} = \argmin_{\gait}\,{\alnth_{\gait}} \ \ |\ \ \fiber_{\gait} = g_c.
    \label{eq:soptgaitdef2}
\end{equation}
This minimization is a classic optimization problem with a well-known solution via Lagrange multipliers (see the Appendix), in which the gradients of the cost function $\alnth_{\gait}$ and the constraint function $\fiber_{\gait}$ are parallel to each other and scaled by an (initially unknown) factor $\lambda$.\footnote{The unconstrained optimum can be seen as a special case of the constrained optimum, in which $\lambda = \fiber_{\gait}/\alnth_{\gait}$.}

To find the optimum, we introduce $\lambda$ as a Lagrange multiplier, and use it to define a Lagrangian function $\mathcal{L}$ over the gait parameters,
\begin{equation}
    \mathcal{L}(p,\lambda) = \alnth_{\gait}(p) - \lambda(\fiber_{\gait}(p)-g_c).
\end{equation}
The parameters of the step-optimal gait can then be found by solving for the point $(p^*,\lambda^*)$ where the derivative of $\mathcal{L}$ with respect to both the gait parameters and the Lagrange multiplier is zero,
\begin{equation}
    \nabla_{p,\lambda} \mathcal{L} = \mathbf{0}.
    \label{eq:constrainedoptflow}
\end{equation}
As in the unconstrained optimization case, the lefthand side of~\eqref{eq:constrainedoptflow} defines a vector field which can be flowed along to its equilibrium to find the solution.

\emph{Example: Weighted Isoareal Problem} The constrained version of the weighted area-perimeter problem is the \emph{weighted isoareal problem}: find the shortest closed curve that encircles a given weighted area. The contours in Fig. \ref{fig:LagrangeExample} represent level sets of the cost function (dashed) and net displacement (solid). The points where these curves touch minimize the cost along the displacement level set, and so also maximize area per perimeter on that level set.

%%%%%%%%%%%%%%%%%%%%%%%%% Optimization %%%%%%%%%%%%%%%%%%%%%%%%%
\subsection{Optimal Locus Generator}
The key idea behind this paper replaces iterative gait optimization for net displacement with solving one system whose solutions represent optimal gait families.
Here, we describe our approach for the optimal locus generator, which is based on the Lagrange multiplier method; additional discussion of Lagrange multipliers may be found in the Appendix.

\begin{figure*}[t]
    \centering
    \includegraphics[width=\linewidth]{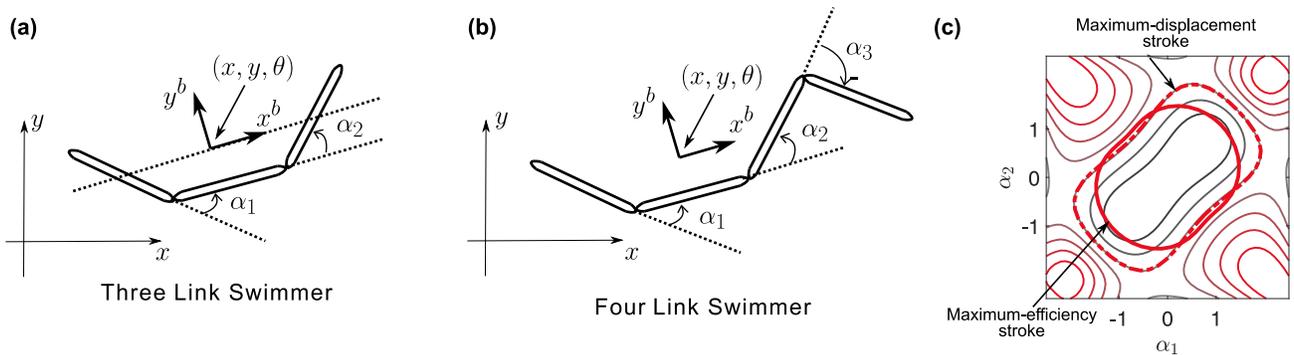}
    \caption{Geometry and shape variables of the example systems. The systems are articulated swimmers. (a) The shape of the three-link swimmer is described by two shape variables. (b) The shape of the four-link swimmer is described by three shape variables. (c) Two optimal gaits of viscous three-link swimmer: Maximum-displacement optimal gait and Maximum-efficiency optimal gait. These two gaits are found in \cite{tam_optimal_2007,ramasamy_geometry_2019}
    }
    \label{fig:SystemExample}
\end{figure*}

Fig. \ref{fig:LagrangeExample} illustrates the geometric intuition underlying the Lagrange multiplier method: the gradient of the cost is parallel to the gradient of the displacement at each stationary point $p^*$, regardless of the level of constraints $g_c$. With respect to $g_c$ as a variable, $p^*$ formulates the trajectory representing a family of optimal gait parameters whose members correspond to each constrained displacement. From the geometric interpretation, we  deduce that the Lagrange multiplier is the value normalizing the gradient of the net displacement with respect to the gradient of the cost at the optimal point $p^*$ \cite{fletcher2013practical,kalman_leveling_2009}. From this, we can $\lambda^*$ as a function of $p$:
\begin{equation}
    \lambda^* = 
    \frac{\nabla_p g^{*T}\cdot\nabla_p s^*}
    {||\nabla_p g^*||^2}.
    \label{eq:OptimalLambda}
\end{equation}

Then, the gradient of the Lagrangian function can be expressed as a function of $p^*$. It is natural to ask: ``What if we choose $\lambda^*$ derived at the non-optimal point when generating the gradient of the Lagrangian?'' The normalizing property of $\lambda^*$ results in the same norm and sign of $\nabla_p g$ as those of $\nabla_p s$. We know that the gradient of the Lagrangian is constructed by the summation of the normalized gradient of the objective function and the constraint. Thus, $\nabla_p \mathcal{L}(p,\lambda^*)$ can be interpreted as the vector pointing in the direction of increasing the objective and the constraint function at the same time. As shown in Fig. \ref{fig:LagrangeExample}, the behavior of the negative gradient vectors of the Lagrangian acts as the optimizer which minimizes the cost with some constraint function. This implies that the norm of the gradient acts as a measure of the optimality of the gait.

Rather than optimizing gaits for all constraints several times, it is desirable to evaluate the trajectory of $p^*$ directly. To derive the solution vector pointing in the direction of from the current optimal parameter to the next optimal parameters, we used the property that the optimal point $p^*$ should be the stationary point of the Lagrangian function satisfying the second derivative test.

The solution vector is directly related to the null space of the Hessian of the Lagrangian function with $\lambda^*$. The eigenvector of the Hessian corresponding to non-zero eigenvalues will point in the direction of either increasing or decreasing the gradient of the Lagrangian. On the other hand, the eigenvector for zero eigenvalues will aim in the direction of not changing the gradient. This implies that $\dot{p}^*$ is in the null space of the Hessian:
\begin{equation}
    \dot{p}^* \subseteq \mathcal{N}\left(\nabla_{p}^2\mathcal{L}\left(p^*\right)\right),
    \label{eq:resVec}
\end{equation}
where $\mathcal{N}$ is a null space. The Hessian matrix of the Lagrangian function is:
\begin{equation}
    \nabla_p^2\mathcal{L}^* = 
    \nabla_p^2 s^*
    -\lambda^*\nabla_p^2 g^*
    -\nabla_{p}g^*\cdot\nabla_{p}\lambda^{*T}
\end{equation}
and
\begin{multline}
    \nabla_{p}\lambda^* = 
    \frac{1}
    {||\nabla_p g^*||^2}\left(\nabla_p^2 g^*\cdot\nabla_p s^*+\nabla_p g^*\cdot\nabla_p^2 s^*\right)\\
    -\frac{2}
    {||\nabla_p g^*||^4}\left(\nabla_p^2 g^*\cdot\nabla_p g^*\right)\left(\nabla_p g^{*T}\cdot\nabla_p s^*\right).
    \label{eq:OptimaldLambda}
\end{multline}

For higher dimensional parameter spaces, the null space of the Lagrangian function contains not only the desired vectors keeping the solution on the stationary point but also \textit{trivial null vectors} that do not change the gait cycle itself. For example, suppose that lower order Fourier coefficients are chosen as gait parameters. It could be possible that many different sets of parameters generate the same gait cycle because the gait cycle is not affected by the same phase shift of coefficients between shape variables at each order of the Fourier coefficient. Furthermore, there might be more than one desired vector if the objective function is osculated with the constraint function. To prevent choosing only these trivial null vectors, $p^*$ is evaluated by projecting the negative gradient of displacement onto the null space so that the gait cycle can be changed at each iteration.
\begin{equation}
    \dot{p}^* = \text{Proj}_{\mathcal{N}}\left(-\nabla_p g\right)
\end{equation}

\section{Swimmer Locomotion as a Weighted Isoareal Problem}
Here, we apply the Lagrangian optimal locus generator to a set of example systems: drag-dominated articulated swimmers.
For viscous swimmers, the displacement resulting from a specific gait is approximately equivalent to the amount of \textit{constraint curvature} enclosed by the gait.
In effect, this is an application of the optimal locus generator to a weighted isoareal problem.

\subsection{Drag-dominated Swimmers}
We focus on the 3- and 4-link drag-dominated (viscous) swimmers~\cite{Purcell:1977,kelly_geometric_1995,ramasamy_geometry_2019} in our analysis.
These systems are shown in Fig. \ref{fig:SystemExample}(a,b).
The positions of the articulated swimmers are the locations and the orientations of their centroids and mean orientation lines, $g=(x,y,\theta)\in SE(2)$.
The shapes of the articulated swimmers are parametrized by their joint angles, $r=(\alpha_1,\alpha_2)$ for the three-link swimmer and $r=(\alpha_1,\alpha_2,\alpha_3)$ for the four-link swimmer.
Displacement is approximately equal to the weighted area enclosed by a gait; area weighting (in the form of constraint curvature), with accompanying example gaits, is shown in Fig. \ref{fig:SystemExample}(c).

\begin{figure*}[ht]
\includegraphics[width=\textwidth]{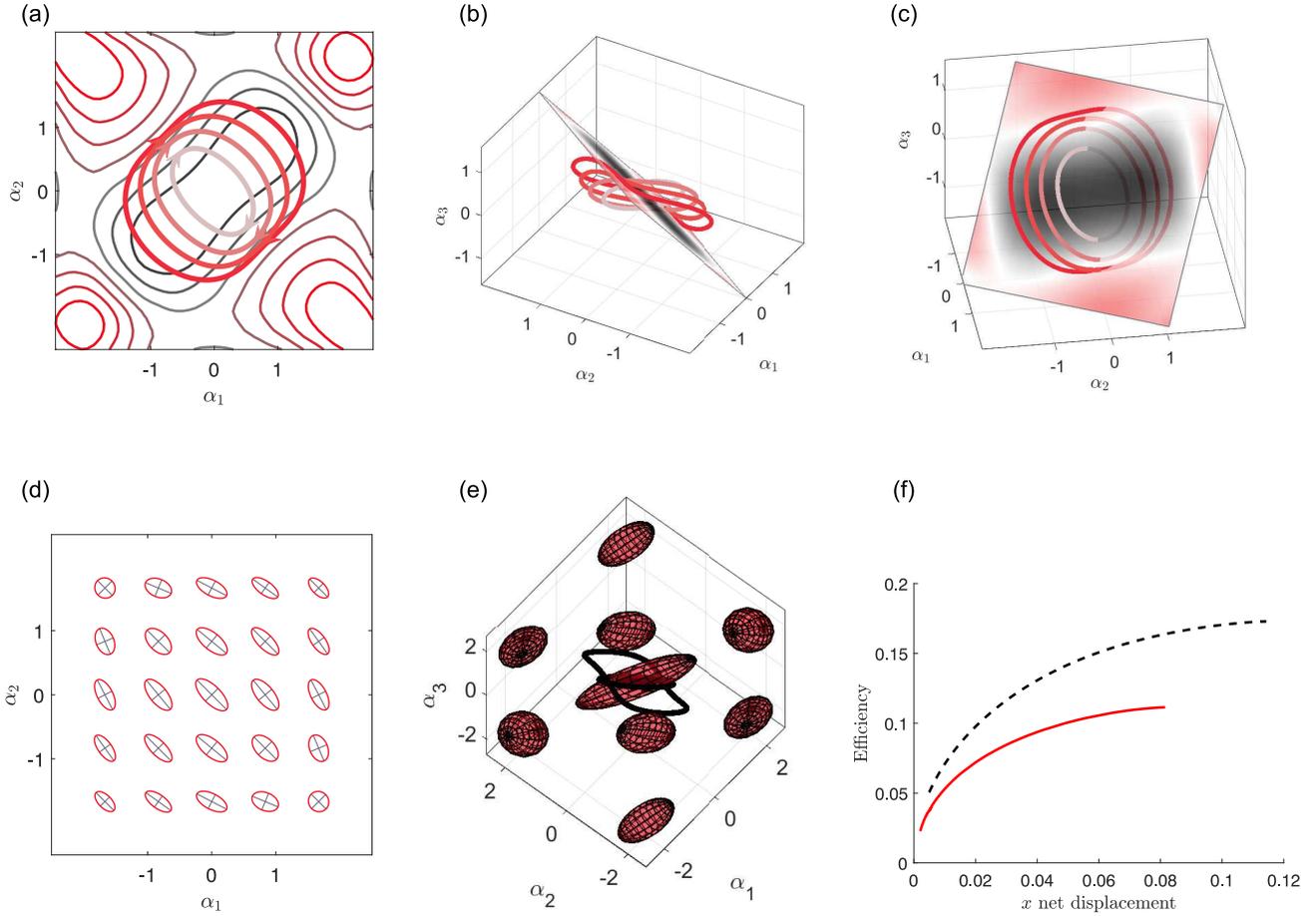}
    \caption{Optimal gaits for viscous three-link and four-link systems. (a) The four step-optimal gaits for the viscous three-link system. (b-c) The four step-optimal gaits for the viscous four-link system. Each line shows the step-optimal gait corresponding each constrained $x$ displacement of the maximum efficiency gait: $1$, $\frac{3}{4}$, $\frac{2}{4}$, $\frac{1}{4}$ of $x$ net displacement of the maximum efficiency gait. (d) The metric ellipse field (Tissot indicatrix) for the three-link system from \cite{hatton_kinematic_2017}, and (e) four-link system. (f) Efficiency gradually decreases as $x$ net displacement decreases. The red solid line represents three-link swimmer, and the black dashed line represents the four-link swimmer.}
    \label{fig:OptGaitResult}
\end{figure*}

\subsection{Application of the Optimal Locus Generator}
Here, we generate optimal gaits for specific amounts of displacement of the viscous swimmers.
We set the initial (seed) gait as the maximum-efficiency gait from \cite{ramasamy_geometry_2019}, as in Fig. \ref{fig:SystemExample}(c).
Gait parameters $p^*$ are derived from this seed gait; the solution vector $\dot{p}^*$ is given by Equation (\ref{eq:resVec}).
In this paper, $\dot{p}^*$ is found numerically with a fourth-order Runge-Kutta method (specifically, MATLAB's \verb|ode45|), using singular value decomposition to find null vectors.
At each iteration, to verify that each evaluated point $p^*$ is a minimizer, we check the norms of the Lagrangian, and perform the second-derivative test (described in the Appendix, Theorem \ref{thm:secondderivative}).
If $p^*$ is not a minimizer or reaches a specified minimum displacement, the integration is stopped.

We applied the optimal locus generator to the viscous three-link swimmer to find the optimal gaits family in the $x$-direction over a single cycle. Each member of the step-optimal gait family induces different step sizes with a range from the net displacement induced by the maximum efficient gait to nearly stopping motion. Four gaits in Fig. \ref{fig:OptGaitResult}(a) represent the step-optimal gaits that generate $1$, $\frac{3}{4}$, $\frac{2}{4}$, $\frac{1}{4}$ of $x$ net displacement of the maximum efficiency gait among the members of the family.  As the constrained net displacement decreases, the curve shape gradually changes to an elliptical shape so that the optimal gait curve contains as much of the constraint curvature as possible while minimizing the path length; the corresponding metric stretch is illustrated in Fig. \ref{fig:OptGaitResult}(d). 

This optimization approach works regardless of the dimension of the shape space, as long as the cost and the constraint functions are well-defined. Fig. \ref{fig:OptGaitResult}(b-c) demonstrates the optimal locus generator results for the viscous four-link system in the $x$-direction. The four gaits in Fig. \ref{fig:OptGaitResult}(b-c) are chosen by the same criteria as them in \ref{fig:OptGaitResult}(a). Because $D(-\mathbf{A})$ can be interpreted as a vector field in three-dimensional shape spaces, every step-optimal gait maximizes the flux of the vector field through a surface bounded by the curve while minimizing the path length. Corresponding metric stretch is illustrated in Fig. \ref{fig:OptGaitResult}(e). 

Fig. \ref{fig:OptGaitResult}(f) shows each system's efficiency at the given net displacement. As expected, the initial, globally optimal gait has maximum efficiency. As net displacement decreases, the efficiency also gradually decreases.

\section{Conclusion}
In this paper, we identified a method for moving across the optimal gaits in the family by expanding the geometric concept of the Lagrange multiplier method; By identifying every possible step-optimal gait for each supplied cost and constraint at once, this method needs less computational cost than optimizing each optimal gaits of the family one at a time. We demonstrated this step-optimization method on two drag-dominated systems: a viscous three-link and four-link swimmer. The optimal locus generator successfully generated the trajectory representing all of the optimal parameters for maximum-efficiency gaits at all displacements.

%The framework we present allows us to understand the constrained optimization problem geometrically. To expect the behavior of the norm of the gradient corresponding to the optimality, the Hessian of the Lagrangian function was considered.

A line of future work is investigating how to expand the optimal locus generator to inertia-dominated systems. The cost for these systems is based on the acceleration during the gait, rather than the path length. This work will require an evaluation of the Hessian of both the cost function (including acceleration) and the net displacement function.

We believe that this work acts as the foundation for new research, including the identification of the efficient steering gait family, by understanding how to optimize with respect to multiple constraints. Furthermore, it is possible to investigate how to capture the most important components of a high-dimensional shape space in an intrinsically two-dimensional structure from the feasibility of the optimal locus generator with shape spaces of any number of dimensions.

\section*{Appendix}

\begin{theorem}[Lagrange Multipliers Theorem]
\label{Thm:Lagrange}
Let there be a set of parameters $p \in \mathbb{R}^n$, $s:\mathbb{R}^n \rightarrow \mathbb{R}$ be a cost/objective function on $p$, and $g:\mathbb{R}^n \rightarrow \mathbb{R}$ be a constraint function on $p$.
We assume that $s$ and $g$ are $C^2$ continuous with respect to $p$; that is, they are continuous and have a first and a second derivative everywhere.

We define $\mathcal{L}:\mathbb{R}^n\times\mathbb{R}\rightarrow\mathbb{R}$ to be a Lagrangian function.\footnote{
    The method of Lagrange multipliers allows us to find critical points of a cost function with respect to constraints. In our case, there is only one constraint equation.
}
Every optimal solution $(p^*,\lambda^*)$ to the optimization problem 
\begin{center}
Minimize $s(p)$ subject to $g(p)=g_c$,
\end{center}
where $g_c$ is the desired value of the constraint function, must satisfy the necessary conditions of optimality\cite{chong2004introduction,kalman_leveling_2009}
\begin{equation}
    \nabla_{p,\lambda}\mathcal{L}(p^*,\lambda^*) = 0,
\label{eq:Lagrange1}
\end{equation}
where
\begin{equation}
    \mathcal{L}(p,\lambda) = s(p) - \lambda(g(p)-g_c)
\label{eq:Lagrange2}
\end{equation}
\qed
\end{theorem}

\begin{theorem}[Second Derivative Test]
\label{thm:secondderivative}
Let $(p^*, \lambda^*)$ be stationary points of the Lagrangian $\mathcal{L}$.
Because its constituent functions are $C^2$ continuous, $\mathcal{L}$ is twice differentiable.
We define the \textit{bordered Hessian}, $\overline{\nabla}^2_p\mathcal{L}$:
\begin{equation}
    \overline{\nabla}^2_p\mathcal{L} =
    \begin{bmatrix}
        0 && 
        \nabla_p g^{*T} 
        \vspace{4pt}\\
        \nabla_p g^*&& 
        \nabla_p^2\mathcal{L^*}
    \end{bmatrix},
    \label{eq:BorderedHessian}
\end{equation}
where $g^*$, $s^*$ are the values of the constraint and cost functions at the stationary point $p^*$, and  $\nabla_p^2\mathcal{L^*}$ is the Hessian of the Lagrangian with respect to $p$ at the stationary point $p^*$\cite{fletcher2013practical}.

The following are true for the stationary points  $(p^*, \lambda^*)$.
\begin{itemize}
    \item If $\overline{\nabla}^2_p\mathcal{L}(p^*, \lambda^*)$ is positive definite, then $p^*$ is a local minimizer subject to $g(p)=g_c$. This is a sufficient optimality condition for our case.
    \item If $\overline{\nabla}^2_p\mathcal{L}(p^*, \lambda^*)$ is negative definite, then $p^*$ is a local maximizer subject to $g(p)=g_c$.
    \item If $\overline{\nabla}^2_p\mathcal{L}(p^*, \lambda^*)$ is indefinite, then $p^*$ is neither a local maximizer nor local minimizer subject to $g(p)=g_c$.
\end{itemize}
\qed
\end{theorem}

%\subsection{Lagrange Multipliers Theorem}

%Note that the stationary points of the Lagrangian function are the only candidates for the optimal point because the condition is necessary but not sufficient. A sufficient condition for minimizers is needed such as the second derivative test. The second derivative of the multi-variable function produces a square matrix (i.e. the Hessian matrix) of size equal to the number of variables. In case of constrained optimization, it is convenient to introduce the bordered Hessian $\overline{\nabla}^2_p\mathcal{L}$ to check the sufficient conditions.
%\begin{equation}
%    \overline{\nabla}^2_p\mathcal{L} =
%    \begin{bmatrix}
%        0 && 
%        \nabla_p g^{*T} 
%        \vspace{4pt}\\
%        \nabla_p g^*&& 
%        \nabla_p^2\mathcal{L^*}
%    \end{bmatrix},
%    \label{eq:BorderedHessian}
%\end{equation}
%where $g^*$ is the net displacement function at the optimal point $p^*$, $s^*$ is the cost function at the optimal point $p^*$ and  $\nabla_p^2\mathcal{L^*}$ is the Hessian matrix of the Lagrangian function with respect to $p$ at the optimal point $p^*$.

\section*{ACKNOWLEDGMENT}
We thank Noah J. Cowan, Siming Deng, and Nathan Justus for many insightful discussions, and the NSF for support via Grants CMMI-1653220.

%%%%%%%%%%%%%%%%%%%%%%%%%%%%%%%%%%%%%%%%%%%%%%%%%%%%%%%%%%%%%%%%%%%%%%%%%%%%%%%%

\bibliographystyle{IEEEtran}
\bibliography{reference}

\end{document}